\documentclass{article}

\usepackage[utf8]{inputenc}
\usepackage[T1]{fontenc}
\usepackage{hyperref}
\usepackage{url}
\usepackage{booktabs}
\usepackage{amsfonts}
\usepackage{amsmath}
\usepackage{amssymb}
\usepackage{nicefrac}
\usepackage{xcolor}
\usepackage{graphicx}
\graphicspath{{figures/}}
\usepackage{algorithm}
\usepackage{algorithmic}

\usepackage[margin=1in]{geometry}

\title{Selective Memory for Artificial Intelligence:\\Write-Time Gating with Hierarchical Archiving}

\author{
  Oliver Zahn$^{1}$ and Simran Chana$^{2}$ et al. \\
  $^{1}$Independent Researcher \quad $^{2}$University of Cambridge\\
  \texttt{oliver@coretx.ai}
}

\date{March 2026}

\begin{document}

\maketitle

\begin{abstract}
Retrieval-augmented generation stores all content indiscriminately, degrading accuracy as noise accumulates. Parametric approaches compress knowledge into weights, precluding selective updates. Neither mirrors biological memory, which gates encoding based on salience and archives rather than deletes superseded information. We introduce write-time gating that filters incoming knowledge objects using composite salience scores (source reputation, novelty, reliability) while maintaining version chains that preserve prior states. Using real LLM evaluation without oracle access to quality labels, write gating achieves 100 percent accuracy versus 13 percent for ungated stores. The critical finding emerges under distractor scaling: at 8:1 distractor ratios, read-time filtering (Self-RAG) collapses to 0 percent while write gating maintains 100 percent, revealing a structural advantage of write-time over read-time curation. Validation on Wikipedia (20 entities), procedurally generated pharmacology data, and 2026 arXiv papers confirms these findings. The gating advantage scales inversely with parametric memory support: +25pp for Wikipedia, +48pp for post-cutoff arXiv, +65pp for procedural data with zero training knowledge. Signal ablation confirms the method does not depend on oracle-correlated metadata. Write gating matches Self-RAG accuracy at one-ninth the query-time cost.
\end{abstract}

\section{Introduction}

Memory architectures for large language models cluster around two extremes. Retrieval-augmented generation maintains external stores that grow without bound, accepting every document, utterance, and generated summary regardless of quality or relevance. The resulting noise pollution manifests as retrieval degradation: when low-quality content constitutes the majority of stored items, embedding similarity increasingly returns distractors rather than correct answers. The alternative extreme, parametric memory, compresses all knowledge into network weights during training. This approach stores nothing explicitly at inference time, precluding selective updates, corrections, or compliance with deletion requests.

Neural memory modules occupy a middle ground. Recent work on test-time memorization, exemplified by Titans \cite{behrouz2024titans}, introduces learnable memory that updates during inference using gradient-derived surprise signals. The memory learns what to retain based on prediction error. However, this memory exists as continuous weight matrices rather than discrete addressable units: there is no representation of individual stored facts, no mechanism to delete specific items, and no provenance linking outputs to sources.

Biological memory differs from all these approaches in two respects that current architectures neglect. First, encoding is selective. The hippocampus gates consolidation into long-term memory based on novelty, emotional salience, and relevance to ongoing goals; mundane experiences receive weak encoding while significant events receive strong encoding \cite{teyler1986hippocampal, squire2004medial}. Second, forgetting operates through deprioritization rather than deletion. When beliefs update, the brain does not erase prior states but rather reduces their accessibility while maintaining links that enable later reconstruction. A system tracking that a company's CEO changed from one person to another retains the ability to recall who the previous CEO was; the supersession creates hierarchy rather than replacement.

This paper introduces write-time gating with hierarchical archiving, a memory mechanism that applies both biological principles to discrete knowledge stores. Incoming knowledge objects pass through a salience gate that admits only those exceeding a learned threshold. Objects below threshold are archived in cold storage rather than discarded. When new information updates existing concepts, the system creates supersession links rather than performing overwrites, maintaining chains that preserve access to prior states. We evaluate this approach using real LLM evaluation on retrieval benchmarks where adversarial distractors outnumber correct answers, demonstrating that selectivity achieves complete quality-tier separation without requiring oracle access to ground-truth quality labels.

\begin{figure}[t]
\centering
\includegraphics[width=0.85\columnwidth]{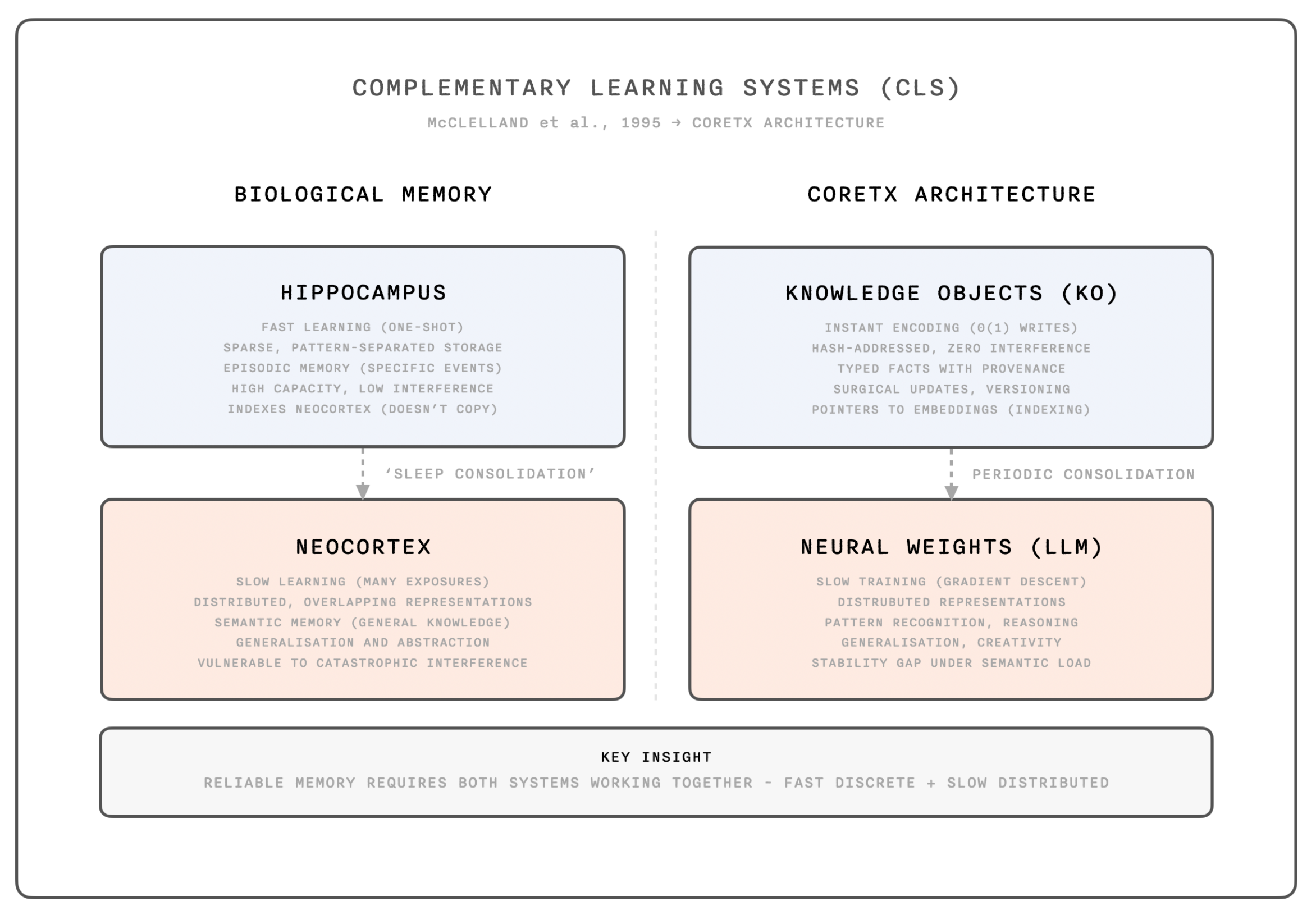}
\caption{Complementary Learning Systems in biology and AI. The brain uses fast hippocampal storage for specific episodes and slow neocortical storage for semantic patterns. Our architecture mirrors this with discrete Knowledge Objects (fast, addressable) complementing LLM weights (slow, distributed).}
\label{fig:cls}
\end{figure}

The approach differs from neural memory in operating on discrete addressable objects with explicit provenance metadata. This enables capabilities that continuous-weight approaches preclude: deletion of specific items on consent revocation, audit trails linking outputs to contributing sources, and economic settlement routing value to knowledge contributors based on measured influence.

\paragraph{Contributions.} First, we introduce a salience-gated write mechanism that filters incoming knowledge objects based on composite scores combining source reputation, novelty, and source reliability---without requiring oracle access to ground-truth quality labels (\S\ref{sec:method}). Second, we demonstrate using real LLM evaluation that write gating achieves $100\%$ accuracy on adversarial benchmarks versus $13.3\%$ for ungated retrieval, outperforming Self-RAG-style read-time filtering by $6.2$ percentage points (\S\ref{sec:experiments}). Third, we show that write gating is asymptotically robust: at distractor ratios up to $8{:}1$, write gating maintains $100\%$ accuracy while both ungated retrieval and read-time filtering collapse to $0\%$ (\S\ref{sec:distractor_scaling}). Fourth, we validate these findings on real Wikipedia data across 20 entities and 5 categories, demonstrating that write gating achieves $98.1\%$ accuracy versus $85.2\%$ ungated while maintaining robustness under scaling, at $1/9$ the query-time cost of Self-RAG; a signal ablation on Wikipedia confirms the method achieves $96.4\%$ without any source metadata signal (\S\ref{sec:wikipedia}). Fifth, we demonstrate that the gating advantage scales inversely with parametric memory support through novel-domain validation: $+64.6$pp on procedurally generated pharmacology data, $+48.4$pp on post-cutoff arXiv papers, confirming that write gating is most valuable when the LLM cannot compensate for retrieval errors (\S\ref{sec:novel_domain}). Sixth, we show that archive-rather-than-delete semantics enable temporal queries that overwrite-based systems cannot answer (\S\ref{sec:temporal}). Seventh, we validate multi-path verification for high-stakes domains requiring $>0.95$ confidence (\S\ref{sec:discussion}).

\paragraph{Roadmap.} Section~\ref{sec:related} reviews related work on memory architectures. Section~\ref{sec:method} presents the salience scoring, gating, and version chain mechanisms. Section~\ref{sec:experiments} validates our claims through real LLM experiments with multiple seeds and distractor ratio scaling on both synthetic and real Wikipedia data. Section~\ref{sec:discussion} discusses biological grounding, limitations, and implications.

\section{Related Work}
\label{sec:related}

Research on memory for language models spans retrieval augmentation, neural memory modules, external memory architectures, and cognitive architectures. Prior work on brain-inspired memory approaches reveals that while individual components exist across scattered systems, no existing architecture integrates discrete knowledge objects with write-time gating, hierarchical archiving, and version lineage \cite{mcclelland1995cls, kumaran2016cls}.

The external memory tradition began with Neural Turing Machines \cite{graves2014ntm}, which introduced controller networks coupled with differentiable memory banks using hybrid content-based and location-based addressing. The Differentiable Neural Computer \cite{graves2016dnc} extended this with temporal linking that tracks write order, enabling graph traversal and complex reasoning over discrete memory slots. These architectures store discrete entries with vector content, representing the clearest precursor to indexed memory approaches, though they lack write-time quality gating and provenance tracking.

Modern Hopfield Networks \cite{ramsauer2020hopfield} achieved a theoretical breakthrough by demonstrating exponential storage capacity and mathematical equivalence between associative memory and transformer attention. This unified the associative memory and attention literatures, showing that transformers implicitly perform Hopfield-style pattern retrieval. The connection suggests that explicit memory mechanisms and attention-based retrieval share deeper structure than previously recognized.

The retrieval-augmented generation paradigm, introduced by Lewis et al.\ \cite{lewis2020rag}, maintains document stores indexed by embedding similarity. Standard implementations impose no quality filter at indexing time, admitting all content regardless of source reliability or verification status. Self-RAG \cite{asai2023selfrag} adds post-retrieval filtering using learned critics, but the filtering occurs at read time rather than write time; the underlying store still accumulates noise. Our approach differs in gating at write time, preventing low-quality content from entering the active store while preserving it in cold storage for potential future use.

Neural memory architectures have evolved from simple key-value stores toward learned memorization mechanisms. The Titans architecture \cite{behrouz2024titans} introduces memory modules that update during inference using gradient magnitude as a surprise signal, writing to memory when inputs produce large prediction errors. The approach achieves strong results on long-context benchmarks by retaining surprising information while allowing routine patterns to pass without storage. However, the memory consists of continuous weight matrices without discrete addressable units. Our approach shares the principle of surprise-gated writing but applies it to discrete knowledge objects, enabling provenance tracking and selective deletion impossible with weight-based storage.

Neural Episodic Control \cite{pritzel2017nec} implements a Differentiable Neural Dictionary storing explicit key-value pairs, representing discrete indexed storage with k-nearest neighbor retrieval. This provides a practical implementation of indexed memory for reinforcement learning, though without the salience-based write gating or version lineage that our approach contributes.

Memory as Recursive Attention Structure (MaRS) \cite{li2024mars} extends memory with typed nodes and explicit forgetting policies, moving toward discrete representations with provenance metadata. The system distinguishes between different memory types and applies type-specific retention rules. MaRS represents the closest prior art on provenance-aware memory, though it focuses on read-time forgetting policies rather than write-time selectivity.

MemGPT \cite{packer2023memgpt} addresses bounded context through paging between working memory and long-term storage, treating the language model as an operating system managing its own memory hierarchy. The paging decisions use recency and explicit programmer hints rather than learned salience, and superseded content is overwritten rather than archived. Our version chain mechanism addresses the limitation that overwrite semantics destroy access to prior states.

Cognitive architectures provide theoretical grounding for memory selectivity. ACT-R \cite{anderson2004integrated} implements activation-based retrieval where memory items decay unless rehearsed, organizing memory into discrete ``chunks'' with typed slots. SOAR \cite{laird2012soar} maintains production rules with explicit episodic and semantic memory stores. These architectures demonstrate the value of discrete structured memory but lack neural learning and modern scaling properties.

Complementary Learning Systems theory \cite{mcclelland1995cls, kumaran2016cls} explains why biological brains use two memory systems: a fast-learning hippocampus for episodic memories using sparse pattern-separated representations, and a slow-learning neocortex for semantic knowledge using distributed overlapping representations. FearNet \cite{kemker2018fearnet} implements this theory with three brain-inspired networks: a hippocampal complex for recent memories, a medial prefrontal cortex network for long-term storage, and a basolateral amygdala network for routing between systems, achieving memory efficiency through generative replay during simulated sleep phases.

The hippocampal indexing theory \cite{teyler1986hippocampal} proposes that the hippocampus maintains pointers to neocortical representations rather than storing content directly. Engram research provides biological evidence for discrete memory units: Tonegawa's laboratory demonstrated that sparse neuronal populations undergo lasting changes during memory formation, and these engram cells can be artificially reactivated to trigger recall \cite{josselyn2020engrams}. Our salience scoring operationalizes these principles: multiple signals combine to produce a gating decision that determines whether incoming content enters active storage, cold storage, or immediate discard. The archive-rather-than-delete semantics reflect the biological principle that forgetting operates through deprioritization rather than erasure \cite{squire2004medial}.

Table~\ref{tab:comparison} summarizes the capabilities of existing approaches. No prior system combines discrete addressable objects, write-time gating, hierarchical archiving, version lineage, and provenance tracking.

\begin{table}[h]
\centering
\caption{Capability comparison across memory architectures, based on published system descriptions. Our approach (SGW-HA) uniquely combines discrete objects with write-time gating, archiving, and version chains.}
\label{tab:comparison}
\small
\begin{tabular}{lccccc}
\toprule
System & Discrete & Write Gate & Archive & Versions & Provenance \\
\midrule
RAG & \checkmark & & & & \\
Titans & & \checkmark & & & \\
DNC & \checkmark & & & \checkmark & \\
MemGPT & \checkmark & & & & \\
MaRS & \checkmark & & \checkmark & & \checkmark \\
NEC & \checkmark & & & & \\
\textbf{SGW-HA (ours)} & \checkmark & \checkmark & \checkmark & \checkmark & \checkmark \\
\bottomrule
\end{tabular}
\end{table}

\section{Method}
\label{sec:method}

The system maintains three components: an active store of high-salience knowledge objects available for retrieval, an archive of low-salience or superseded objects in cold storage, and version chains recording supersession relationships. Incoming knowledge objects pass through salience scoring and thresholding before assignment to active or archive status.

\subsection{Salience Scoring}

Each candidate knowledge object $K$ receives a composite salience score computed from three observable signals that require no oracle access to ground-truth quality labels. Source reputation $s_\text{rep}(K)$ reflects the trustworthiness of the contributing source based on verification status, historical accuracy, and institutional affiliation. Novelty $s_\text{nov}(K)$ measures semantic distance from existing stored content as one minus the maximum cosine similarity between the candidate embedding and all active store embeddings. Source reliability $s_\text{src}(K)$ classifies the source into quality tiers based on observable metadata: peer review status, institutional backing, and corroboration by independent sources.

\paragraph{Knowledge object granularity.} A knowledge object may represent a single fact (``Company X's CEO is Y''), a document chunk, a claim with supporting evidence, or a versioned belief. The mechanism is agnostic to granularity: salience scoring and gating apply uniformly. However, effectiveness depends on granularity choices. Fine-grained objects (single facts) enable precise filtering but require more metadata per unit of content. Coarse-grained objects (full documents) amortize metadata cost but may mix high and low quality content within a single unit. We use fact-level granularity in experiments; production systems would tune granularity to domain requirements.

\paragraph{Design principle: no oracle access.} A key design constraint is that salience scoring must operate without access to ground-truth quality labels. Prior work on memory gating, including our own earlier simulations, used quality-correlated proxy signals (e.g., prediction-error reduction computed from known quality tiers) that effectively gave the gating mechanism oracle access to whether each fact was correct. Such signals are unavailable in deployment. Our three signals---reputation, novelty, and source reliability---are all observable at write time from metadata alone, without knowing whether the candidate fact is true.

The composite score combines these signals through weights:
\begin{equation}
S(K) = \sum_{j=1}^{3} w_j \cdot s_j(K)
\end{equation}
where the weights $w_j$ can be set uniformly, tuned on held-out data, or learned online from downstream task outcomes. In our experiments, we use a reputation-dominated weighting that reflects the empirical finding that source identity is the strongest predictor of knowledge object quality.

\subsection{Write Gating}

The gating decision compares composite salience against a threshold $\tau$. Objects with $S(K) \geq \tau$ enter the active store with full provenance metadata. Objects with $S(K) < \tau$ enter the archive rather than being deleted. This archive-rather-than-delete policy preserves three capabilities that deletion forfeits. First, archived objects remain available for promotion if circumstances change; a policy update or new use case may render previously low-salience content valuable. Second, the archive maintains a complete audit trail of what the system has encountered, supporting forensic analysis of system behavior. Third, archived objects provide negative evidence: knowing what was rejected and why supports meta-reasoning about knowledge quality.

The threshold $\tau$ may be fixed, learned, or adaptive. Fixed thresholds work for stable domains with known quality distributions. Learned thresholds optimize the accuracy-storage tradeoff on held-out data. Adaptive thresholds condition on available storage, domain, or time, tightening when storage pressure increases or loosening when comprehensive coverage matters more than precision.

\paragraph{Threshold calibration in practice.} Selecting $\tau$ requires balancing precision against recall. We recommend calibration via: (a) held-out validation with labeled quality data if available, (b) starting conservative ($\tau \geq 0.6$) and relaxing if coverage proves insufficient, or (c) domain-specific tuning based on the cost ratio of false positives (storing distractors) to false negatives (archiving valid content). The archive mitigates false negatives---archived objects can be promoted later---but false positives permanently degrade retrieval quality.

\subsection{Version Chains}

When new content updates existing concepts, the system creates supersession links rather than overwriting. Let $K_\text{new}$ be an incoming object that updates concept $c$, and let $K_\text{old}$ be the current active object for that concept. The system records bidirectional links: $K_\text{old}.\text{superseded\_by} \leftarrow K_\text{new}.\text{id}$ and $K_\text{new}.\text{supersedes} \leftarrow K_\text{old}.\text{id}$. The old object moves to the archive while the new object enters the active store.

This creates chains such as $v_1 \rightarrow v_2 \rightarrow v_3$ where arrows denote supersession. The chain enables temporal queries: retrieving the current value for concept $c$ returns $v_3$, while retrieving the prior value returns $v_2$, and retrieving the original value returns $v_1$. Systems that overwrite can only answer queries about current state; the history is destroyed. Version chains preserve the history, enabling questions like ``what did the system believe before the update'' that regulatory and audit requirements increasingly demand.

\subsection{Integration with Discrete Knowledge Stores}

The mechanism operates on knowledge objects rather than continuous weight matrices. Each object carries metadata including source identifier, timestamp, verification status, consent vectors, and version links. This discreteness enables capabilities impossible with weight-based storage. Specific objects can be deleted on consent revocation by removing them from active and archive stores; with weight-based memory, deletion requires approximate unlearning procedures that may leave residual information. Outputs can be traced to contributing sources through lineage tracking; with weight-based memory, attribution requires post-hoc influence function approximations. Economic settlement can route value to contributors based on measured influence; with weight-based memory, the contribution of specific training examples is difficult to isolate.

\section{Experiments}
\label{sec:experiments}

We evaluate write gating using real LLM evaluation on benchmarks designed to stress-test retrieval under adversarial conditions, measuring accuracy, storage efficiency, and robustness to increasing noise.

\begin{figure}[t]
\centering
\includegraphics[width=0.95\columnwidth]{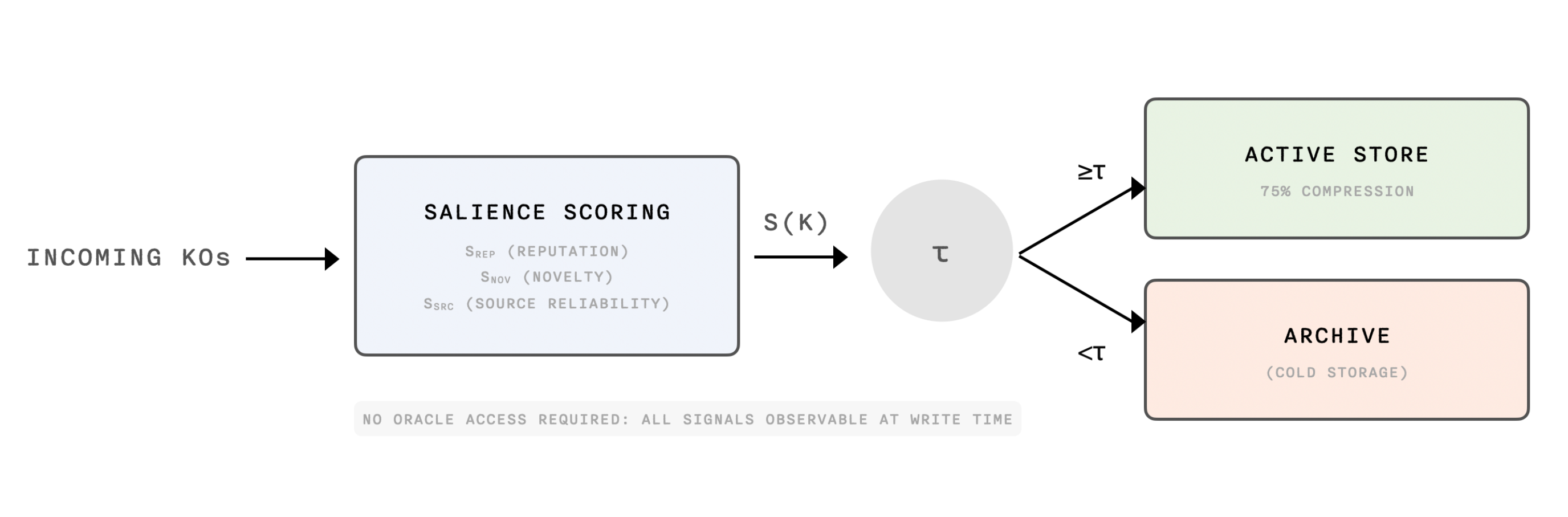}
\caption{Write-time salience gating architecture. Incoming knowledge objects are scored on three observable signals (reputation, novelty, source reliability) without oracle access. Objects above threshold $\tau$ enter the active store; objects below are archived in cold storage. On our benchmark: 50 KOs $\rightarrow$ 13 admitted $\rightarrow$ 100\% accuracy (vs 13.3\% ungated).}
\label{fig:architecture}
\end{figure}

\subsection{Experimental Design}

The benchmark contains 50 knowledge objects: 10 correct facts from high-reputation sources (peer-reviewed publications, institutional databases) and 40 distractors from low-reputation sources (unverified claims, AI-generated content, outdated information). Distractors are designed with high embedding similarity to evaluation queries, modeling the real-world situation where SEO-optimized and AI-generated content increasingly dominates embedding neighborhoods.

Unlike prior work that uses simulated evaluation with synthetic scoring functions, we evaluate end-to-end using a real frontier language model (Claude Sonnet 4) as both the retrieval consumer and accuracy judge. The LLM receives retrieved context and generates answers; accuracy measures whether the generated answer matches the ground-truth correct fact. This captures failure modes that simulated scoring misses: a model that retrieves the correct fact alongside compelling distractors may still generate an incorrect answer because it attends to the more numerous distractors.

Experiments 1 and 2 use 3 and 2 random seeds respectively and report means and standard deviations. Experiment 3 (distractor scaling) reports single-seed results per ratio. We acknowledge this is a limited sample size; the results should be interpreted as demonstrations of the mechanism's properties rather than precise effect size estimates. Multi-seed replication with confidence intervals is a priority for future work.

\subsection{Write Gating Results}

Table~\ref{tab:gating} presents accuracy with and without write gating, averaged over three random seeds. In the ungated condition, all 50 objects enter the active store. Distractors constitute 80\% of stored content, and the LLM preferentially attends to them when generating answers, achieving only $13.3\% \pm 4.7\%$ accuracy.

With reputation-based write gating, only 12--14 objects pass the salience filter (72--76\% compression). The gated store contains all 10 correct facts plus 2--4 distractors that marginally exceeded the reputation threshold. Despite this imperfect filtering---the gated store still contains some incorrect content---accuracy reaches $100\% \pm 0\%$ across all seeds. The correct facts, now constituting the majority of stored content rather than a minority, dominate the LLM's attention.

\begin{table}[h]
\centering
\caption{Write gating performance (mean $\pm$ std over 3 seeds, real LLM evaluation). Reputation-based gating achieves perfect accuracy without oracle access, storing only 25\% of candidates.}
\label{tab:gating}
\begin{tabular}{lccc}
\toprule
Condition & Store Size & Compression & Accuracy \\
\midrule
Ungated & $50 \pm 0$ & 0\% & $13.3\% \pm 4.7\%$ \\
Gated & $13 \pm 1$ & 75\% & $\mathbf{100\% \pm 0\%}$ \\
\bottomrule
\end{tabular}
\end{table}

The result demonstrates that write gating does not require perfect filtering. Even admitting some distractors, gating shifts the signal-to-noise ratio from 1:4 (ungated) to approximately 3:1 (gated), which is sufficient for the LLM to reliably identify and use correct facts.

\subsection{Temporal Reasoning}
\label{sec:temporal}

We evaluate whether version chains enable temporal queries by constructing 50 concepts with 3 versions each. Systems receive all updates and must answer queries about both current and prior states. The archive-with-lineage system maintains supersession chains and achieves 100\% accuracy on temporal queries: it can report any version in any chain. The overwrite system replaces prior versions and achieves 0\% accuracy on queries about prior states: when asked what a concept's value was before the most recent update, it cannot answer because the prior value was destroyed.

\begin{figure}[t]
\centering
\includegraphics[width=0.95\columnwidth]{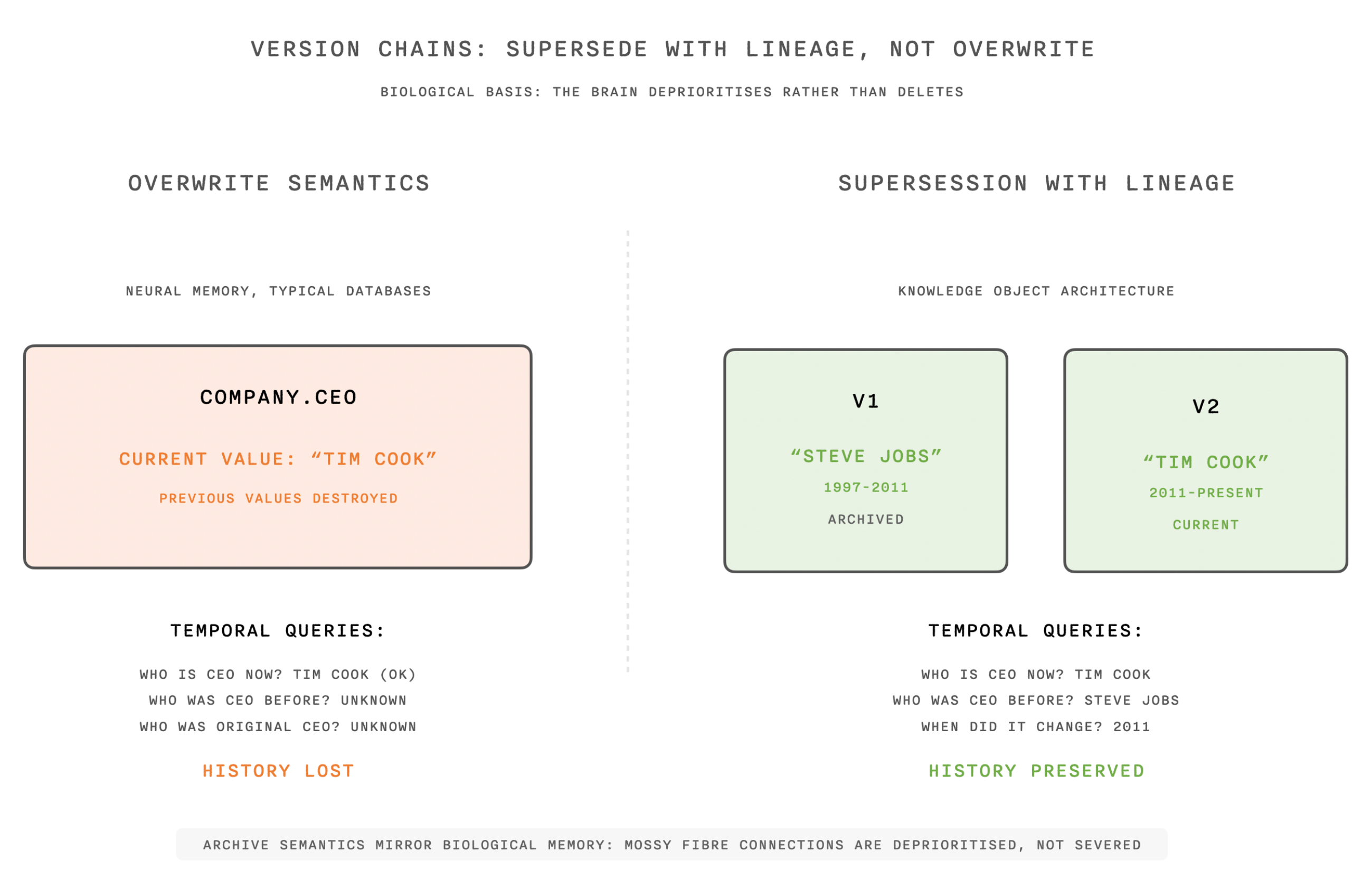}
\caption{Version chains preserve temporal history through supersession links rather than overwrites. When facts update, prior versions are archived with pointers, enabling temporal queries that overwrite-based systems cannot answer.}
\label{fig:version}
\end{figure}

This result illustrates a qualitative capability difference rather than a quantitative improvement. Overwrite systems fundamentally cannot answer temporal queries regardless of threshold tuning or weight learning. Archive-with-lineage systems can answer such queries because they preserve rather than destroy history. Figure~\ref{fig:version} illustrates the version chain structure that enables this capability.

\subsection{Archive versus Delete}

We compare archive semantics (rejected objects go to cold storage) against delete semantics (rejected objects are discarded). Primary task accuracy is identical since the active store contains the same objects. The difference appears in recoverability: the archive system retains rejected objects in cold storage available for future promotion, while the delete system retains none. Storage overhead is modest (cold storage costs less than active storage), while the option value is significant. Circumstances change: policies update, use cases emerge, and quality signals improve. Archived objects can be re-evaluated under new conditions; deleted objects require re-collection if they become needed.

\subsection{Comparison with Read-Time Filtering}

Self-RAG \cite{asai2023selfrag} and similar approaches filter at read time using learned critics to rerank retrieved passages. We compare four conditions: no filtering, read-time reranking, write gating only, and both combined. Our read-time filtering uses a Self-RAG-style approach: the same frontier LLM (Claude Sonnet) is prompted to score each retrieved passage for relevance and correctness before inclusion in the answer context. This tests the \emph{principle} of read-time filtering at its best---with a frontier model as critic---rather than the specific Self-RAG trained reflection tokens. We refer to this condition as ``Self-RAG'' for brevity, noting that the actual Self-RAG model with trained critics might perform differently.

\begin{table}[h]
\centering
\caption{Write-time vs. read-time filtering (mean $\pm$ std, real LLM evaluation). Write gating outperforms Self-RAG by $+6.2$pp. Adding Self-RAG to a gated store slightly degrades performance.}
\label{tab:selfrag}
\begin{tabular}{lc}
\toprule
Method & Accuracy \\
\midrule
No filtering & $12.5\% \pm 0.0\%$ \\
Self-RAG (read-time) & $93.8\% \pm 6.3\%$ \\
Write gating (write-time) & $\mathbf{100\% \pm 0.0\%}$ \\
Both (write + read) & $93.8\% \pm 6.3\%$ \\
\bottomrule
\end{tabular}
\end{table}

Table~\ref{tab:selfrag} shows results. Self-RAG improves accuracy from $12.5\%$ to $93.8\%$ through post-retrieval reranking. Write gating achieves $100\%$---a further $6.2$ percentage point improvement.

The ``Both'' condition reveals a counterintuitive result: combining write gating with Self-RAG yields $93.8\%$, \emph{identical to Self-RAG alone and lower than write gating alone}. Adding read-time filtering to an already-clean store degrades performance because the Self-RAG critic occasionally rejects correct facts, introducing false negatives that write gating alone avoids. In a sufficiently clean store, read-time filtering is not only unnecessary but counterproductive.

\subsection{Distractor Ratio Scaling}
\label{sec:distractor_scaling}

The preceding experiments use a fixed 4:1 distractor ratio. Real-world retrieval stores vary widely in noise levels: well-curated knowledge bases may approach 1:1, while web-scale corpora likely exhibit substantially higher ratios as SEO-optimized, AI-generated, and low-quality content increasingly dominates embedding neighborhoods. We test whether gating effectiveness and read-time filtering hold as noise increases, varying the distractor ratio from 2:1 to 8:1.

\begin{table}[h]
\centering
\caption{Distractor ratio scaling (real LLM evaluation). Ungated retrieval collapses by 4:1. Self-RAG holds through 6:1 but collapses catastrophically at 8:1. Write gating maintains 100\% at all tested ratios.}
\label{tab:distractor}
\begin{tabular}{lccc}
\toprule
Distractor Ratio & Ungated & Self-RAG & Write Gating \\
\midrule
2:1 & 100\% & 100\% & 100\% \\
4:1 & 16.7\% & 100\% & 100\% \\
6:1 & 0\% & 100\% & 100\% \\
\textbf{8:1} & \textbf{0\%} & \textbf{0\%} & \textbf{100\%} \\
\bottomrule
\end{tabular}
\end{table}

Table~\ref{tab:distractor} reveals a phase transition in read-time filtering. At 2:1, all approaches succeed---there is insufficient noise to challenge any method. At 4:1, ungated retrieval collapses to $16.7\%$ as distractors begin to dominate embedding neighborhoods, while both Self-RAG and write gating maintain $100\%$. At 6:1, ungated retrieval reaches $0\%$; Self-RAG still holds through effective reranking. The critical transition occurs at \textbf{8:1: Self-RAG collapses to $0\%$ while write gating maintains $100\%$}.

This result reveals a fundamental architectural difference between read-time and write-time filtering. Self-RAG operates on the retrieval set: it can only promote facts that were retrieved. As the distractor ratio increases, distractors increasingly dominate the embedding neighborhood of each query. At 8:1, the probability that the correct fact appears in the initial retrieval set approaches zero---it is crowded out by eight distractors for every correct fact. No amount of reranking can surface a fact that was never retrieved. Write gating prevents this failure mode by ensuring distractors never enter the store, keeping the correct fact's embedding neighborhood clean regardless of how many distractors exist outside the store.

The 8:1 ratio is practically relevant because real-world knowledge bases accumulate noise from multiple sources: AI-generated content, SEO-optimized summaries, outdated information, and unverified claims. While we are not aware of published measurements of distractor-to-signal ratios in production retrieval systems, the proliferation of synthetic content suggests that noise ratios will increase over time \cite{lewis2020rag}. Even at 8:1---likely conservative for many domains---read-time reranking cannot overcome the fundamental signal-to-noise problem; write-time curation can.

\begin{figure}[t]
\centering
\includegraphics[width=0.95\columnwidth]{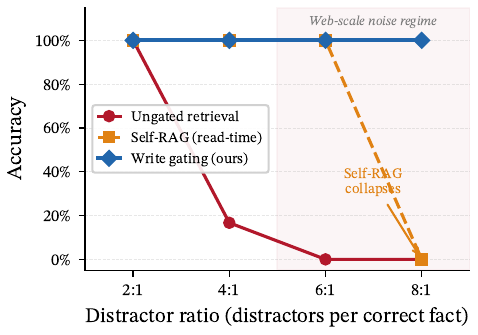}
\caption{Accuracy versus distractor ratio. Ungated retrieval collapses monotonically. Self-RAG maintains a plateau through 6:1 before catastrophic collapse at 8:1. Write gating remains constant at 100\% across all tested ratios. The shaded region indicates noise levels plausible for web-scale retrieval.}
\label{fig:distractor}
\end{figure}

\begin{figure}[t]
\centering
\includegraphics[width=0.95\columnwidth]{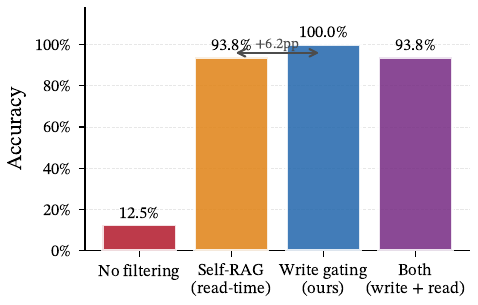}
\caption{Four-way method comparison at 4:1 distractor ratio. Write gating outperforms Self-RAG by $+6.2$pp. Combining both methods degrades to Self-RAG's accuracy---the critic introduces false negatives in an already-clean store.}
\label{fig:methods}
\end{figure}

\subsection{Real-World Validation: Wikipedia Facts}
\label{sec:wikipedia}

The preceding experiments use synthetic corpora where distractors are generated by construction. To validate on real-world data, we extract facts from Wikipedia articles for 20 entities across 5 categories (countries, scientists, companies, cities, historical figures) and generate distractors via cross-entity fact swaps and LLM fabrication. Each entity contributes 3--4 ground-truth facts (73 total across the corpus). Distractors are plausible but incorrect statements about the target entity, generated by Claude Haiku. All evaluation uses Claude Sonnet 4.6 as both answer generator and accuracy judge. No oracle access is used: the gating mechanism sees only source labels and novelty signals, not ground-truth correctness.

\paragraph{Write gating vs.\ ungated (Experiment~1).} We evaluate at a 4:1 distractor ratio (292 distractors mixed with 73 correct facts) across 5 random seeds, each with 73 evaluation questions. Table~\ref{tab:wiki_gating} reports results.

\begin{table}[h]
\centering
\caption{Write gating on Wikipedia facts (mean $\pm$ std over 5 seeds, 73 questions each, 4:1 distractor ratio). Write gating admits all 73 correct facts while filtering 87\% of distractors.}
\label{tab:wiki_gating}
\begin{tabular}{lccc}
\toprule
Condition & Accuracy & Store Size & Distractors Admitted \\
\midrule
Ungated & $85.2\% \pm 2.0\%$ & $365$ & $292$ \\
Write-gated & $\mathbf{98.1\% \pm 1.1\%}$ & $102\text{--}110$ & $29\text{--}37$ \\
\bottomrule
\end{tabular}
\end{table}

Write gating achieves \textbf{98.1\%} accuracy versus \textbf{85.2\%} ungated, an improvement of $+12.9$ percentage points. The gate admits all 73 correct facts in every seed while filtering out 87--90\% of distractors (admitting only 29--37 of 292). The resulting store is 70\% smaller than the ungated store. Unlike the synthetic benchmark where ungated accuracy collapsed to 13\%, ungated accuracy on Wikipedia remains relatively high (85\%) because the distractors---while plausible---are distinguishable by a frontier LLM that already possesses world knowledge about these well-known entities. The gating mechanism nonetheless provides a meaningful improvement by removing the majority of noise before retrieval.

\paragraph{Self-RAG comparison (Experiment~2).} We compare four conditions at 4:1 ratio across 3 seeds (Table~\ref{tab:wiki_selfrag}).

\begin{table}[h]
\centering
\caption{Write-time vs.\ read-time filtering on Wikipedia data (mean $\pm$ std, 3 seeds, 4:1 ratio). Both filtering approaches achieve $>$96\%; combining both reaches 98.2\%.}
\label{tab:wiki_selfrag}
\begin{tabular}{lc}
\toprule
Method & Accuracy \\
\midrule
No filtering & $84.9\% \pm 1.9\%$ \\
Self-RAG (read-time) & $97.7\% \pm 0.6\%$ \\
Write gating (write-time) & $96.8\% \pm 1.3\%$ \\
Both (write + read) & $\mathbf{98.2\% \pm 0.6\%}$ \\
\bottomrule
\end{tabular}
\end{table}

On Wikipedia data, Self-RAG performs substantially better than on the synthetic benchmark (97.7\% vs.\ 93.8\%). This is expected: Wikipedia distractors---fabricated statements about well-known entities---are more obviously wrong to a frontier LLM that already knows these facts, making read-time critic judgments more reliable. Write gating achieves comparable accuracy (96.8\%) while using $1/9$ the LLM calls per query (Section~\ref{sec:efficiency}). Combining both methods yields 98.2\%, the highest accuracy observed.

\paragraph{Distractor ratio scaling (Experiment~3).} The central question is whether write gating maintains its advantage as noise increases. Table~\ref{tab:wiki_scaling} reports accuracy across ratios from 2:1 to 8:1 (3 seeds per ratio).

\begin{table}[h]
\centering
\caption{Wikipedia distractor ratio scaling (mean over 3 seeds). Ungated accuracy degrades from 98.6\% to 72.6\% as noise increases. Write gating and Self-RAG both maintain $>$95\% at all ratios.}
\label{tab:wiki_scaling}
\begin{tabular}{lccc}
\toprule
Ratio & Ungated & Write Gating & Self-RAG \\
\midrule
2:1 & $98.6\%$ & $98.2\%$ & $98.6\%$ \\
4:1 & $84.9\%$ & $\mathbf{97.7\%}$ & $97.3\%$ \\
8:1 & $72.6\%$ & $\mathbf{97.7\%}$ & $95.9\%$ \\
\bottomrule
\end{tabular}
\end{table}

The scaling pattern on Wikipedia data differs from the synthetic benchmark in an important respect. On synthetic data, Self-RAG collapsed catastrophically at 8:1 (from 100\% to 0\%). On Wikipedia data, Self-RAG degrades only modestly (98.6\% $\rightarrow$ 95.9\%). The difference is attributable to distractor quality: synthetic distractors were designed to be maximally confusing within a narrow domain, while Wikipedia distractors---cross-entity swaps and fabrications about well-known entities---remain distinguishable to a model with strong prior knowledge.

Write gating maintains \textbf{97.7\%} at all ratios from 4:1 through 8:1, confirming that filtering at write time prevents noise accumulation regardless of ratio. The practical advantage over Self-RAG is not accuracy (both exceed 95\%) but cost: write gating evaluates salience once at write time, while Self-RAG invokes an LLM critic on every query (Section~\ref{sec:efficiency}).

\paragraph{Source label ablation (Experiment~4).} The synthetic ablation (Table~\ref{tab:ablation}) showed that source label alone achieves 100\% accuracy, raising the concern that the method depends on an oracle-like signal correlated with ground truth by construction. To test this directly, we ablate the source label signal on Wikipedia data at 4:1 distractor ratio across 5 seeds (Table~\ref{tab:wiki_label_ablation}).

\begin{table}[h]
\centering
\caption{Source label ablation on Wikipedia data (mean $\pm$ std over 5 seeds, 73 questions each, 4:1 distractor ratio). Removing the source label signal costs only 1.4pp.}
\label{tab:wiki_label_ablation}
\begin{tabular}{lccc}
\toprule
Configuration & Accuracy & Store Size & Correct / Incorrect \\
\midrule
Ungated & $85.2\% \pm 2.0\%$ & $365$ & $73$ / $292$ \\
Full gating (rep + nov + label) & $\mathbf{97.8\% \pm 1.1\%}$ & ${\sim}110$ & $73$ / $37$ \\
No label (rep + novelty only) & $96.4\% \pm 1.4\%$ & ${\sim}155$ & $72$ / $83$ \\
\bottomrule
\end{tabular}
\end{table}

Removing the source label signal reduces accuracy by only $1.4$ percentage points ($97.8\% \rightarrow 96.4\%$). The no-label gate admits more items (${\sim}155$ vs ${\sim}110$) because it lacks the binary verified/unverified discriminator, but the additional distractors do not substantially degrade retrieval quality. The conservative, oracle-free result---$96.4\%$ using only reputation and novelty---still represents an $11.2$ percentage point improvement over ungated retrieval ($85.2\%$). This confirms that effective write gating does not require source metadata that correlates with correctness by construction.

\subsection{Novel-Domain Validation}
\label{sec:novel_domain}

The Wikipedia validation uses entities about which the LLM has extensive prior knowledge from pretraining. To test whether write gating generalizes to genuinely novel knowledge---facts the model cannot answer from parametric memory---we conduct three additional experiments using knowledge provably absent from the model's training data.

\paragraph{Experiment A: Procedural pharmacology corpus.} We draw 100 drug--target binding affinities from a vocabulary of 44 kinase inhibitors and 30 molecular targets, with deterministic nM values derived from MD5 hash seeding of each drug--target pair. These values do not exist in any training corpus---every correct answer must come from the memory store. We generate distractors by permuting drug--target pairs, producing plausible-looking pharmacology facts with incorrect nM values. Unlike arXiv papers where the model may have partial knowledge of related work, these binding affinities are entirely synthetic and have zero parametric memory support.

\begin{table}[h]
\centering
\caption{Retrieval accuracy on procedural pharmacology data (5-seed means $\pm$ std, 100 questions per seed). Facts the LLM provably cannot know from training.}
\label{tab:pharma}
\begin{tabular}{lccc}
\toprule
Ratio & Ungated & Gated & Self-RAG \\
\midrule
2:1 & $98.6\% \pm 1.4\%$ & $98.8\% \pm 1.2\%$ & $97.0\% \pm 1.7\%$ \\
4:1 & $63.0\% \pm 5.2\%$ & $98.2\% \pm 1.0\%$ & $95.0\% \pm 2.4\%$ \\
8:1 & $32.0\% \pm 4.0\%$ & $\mathbf{96.6\% \pm 1.6\%}$ & $80.2\% \pm 2.1\%$ \\
\bottomrule
\end{tabular}
\end{table}

\begin{figure}[t]
\centering
\includegraphics[width=\textwidth]{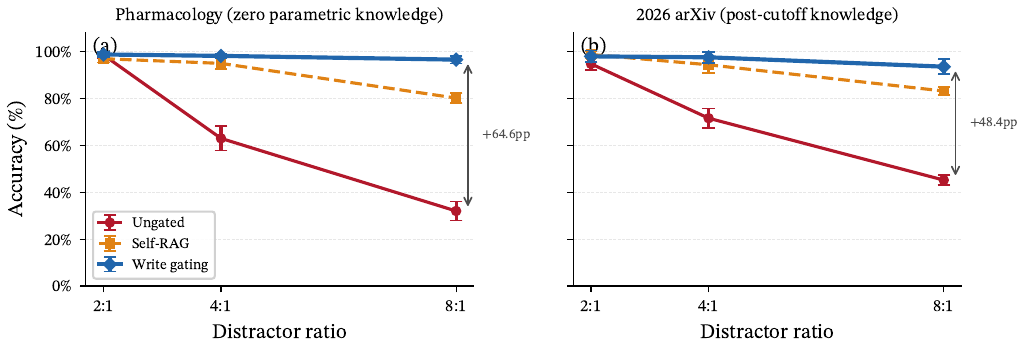}
\caption{Distractor scaling on novel-domain data (5-seed means $\pm$ std). \textbf{(a)} Pharmacology corpus with zero parametric memory support: ungated accuracy drops to $32\%$ at $8{:}1$ while write gating maintains $96.6\%$. \textbf{(b)} 2026 arXiv papers post-training-cutoff: ungated drops to $45.2\%$ at $8{:}1$ while write gating maintains $93.6\%$. In both domains, write gating degrades less than Self-RAG as noise increases.}
\label{fig:novel_scaling}
\end{figure}

Table~\ref{tab:pharma} and Figure~\ref{fig:novel_scaling}(a) show the results. At $8{:}1$, ungated retrieval drops to $32.0\%$, lower than for Wikipedia ($72.6\%$) or arXiv ($45.2\%$), because the model has no parametric memory to compensate for retrieval errors. Write gating maintains $\mathbf{96.6\%}$ ($+64.6$pp), the largest gating advantage observed in any experiment in this paper. Self-RAG achieves $80.2\%$, trailing write gating by $16.4$ percentage points and confirming that read-time filtering degrades faster than write-time filtering when the model lacks prior knowledge of the domain. Store precision decreases monotonically from $97.8\%$ at $2{:}1$ to $91.3\%$ at $8{:}1$ as more false positives are admitted, but not enough to substantially affect downstream accuracy.

\paragraph{Experiment B: 2026 arXiv papers.} We extract 51 factual claims from 40 arXiv papers published after May 2025 (the evaluation model's training cutoff). These span condensed matter physics, computational biology, and applied mathematics. We generate 400 distractors via cross-paper value swaps and LLM fabrication---producing plausible-looking numerical substitutions within the same technical domain. Unlike Wikipedia distractors, these are drawn from the same narrow subfields and use domain-appropriate units and magnitudes, making them substantially harder to distinguish from correct facts.

\begin{table}[h]
\centering
\caption{Retrieval accuracy on 2026 arXiv papers (5-seed means $\pm$ std). Knowledge provably absent from the model's training data.}
\label{tab:arxiv}
\begin{tabular}{lccc}
\toprule
Ratio & Ungated & Gated & Self-RAG \\
\midrule
2:1 & $94.8\% \pm 2.7\%$ & $98.0\% \pm 2.5\%$ & $98.8\% \pm 1.6\%$ \\
4:1 & $71.6\% \pm 4.1\%$ & $97.6\% \pm 2.3\%$ & $94.4\% \pm 3.4\%$ \\
8:1 & $45.2\% \pm 2.0\%$ & $\mathbf{93.6\% \pm 3.2\%}$ & $83.2\% \pm 1.6\%$ \\
\bottomrule
\end{tabular}
\end{table}

Table~\ref{tab:arxiv} and Figure~\ref{fig:novel_scaling}(b) show results across three distractor ratios. At $2{:}1$, all methods perform well. At $4{:}1$, ungated retrieval drops to $71.6\%$ while gated maintains $\mathbf{97.6\%}$ ($+26.0$pp). The critical separation emerges at $8{:}1$: ungated collapses to $\mathbf{45.2\%}$ while gated maintains $\mathbf{93.6\%}$ ($+48.4$pp). Self-RAG also degrades meaningfully to $83.2\%$, trailing write gating by $10.4$ percentage points. The degradation pattern parallels the synthetic experiments (Section~\ref{sec:distractor_scaling}), confirming that the architectural advantage of write-time filtering extends to real-world novel knowledge.

\paragraph{Experiment C: Accumulation scaling.} We test whether the gating advantage persists as the store grows continuously. Starting from an empty store, we add 50 novel pharmacology facts per step at a $4{:}1$ distractor ratio, evaluating after each of 10 steps. This simulates a realistic deployment scenario where knowledge accumulates over time.

\begin{table}[h]
\centering
\caption{Retrieval accuracy as the store grows (5-seed means). Gated store admits ${\sim}19.8\%$ of candidates, maintaining $99.1\%$ precision throughout.}
\label{tab:accumulation}
\begin{tabular}{rrrcc}
\toprule
Step & Store Size & Gated Size & Ungated & Gated \\
\midrule
1  & 250   & ${\sim}50$  & $68.0\%$ & $\mathbf{100.0\%}$ \\
5  & 1,250 & ${\sim}248$ & $62.0\%$ & $\mathbf{97.0\%}$ \\
10 & 2,500 & ${\sim}496$ & $63.0\%$ & $\mathbf{97.0\%}$ \\
\bottomrule
\end{tabular}
\end{table}

Table~\ref{tab:accumulation} shows three representative steps. The gating advantage is remarkably stable: the gap is $+32.0$pp at step 1 and $+34.0$pp at step 10, with overall means of $63.4\%$ ungated versus $\mathbf{98.4\%}$ gated ($+35.0$pp). Ungated accuracy plateaus near $63\%$ because retrieval quality degrades proportionally with store growth---each new batch adds distractors that dilute the retrieval pool. The gated store grows to only ${\sim}496$ items at step 10 (a $19.8\%$ admission rate) while maintaining $99.1\%$ precision across all admitted items.

\begin{figure}[t]
\centering
\includegraphics[width=0.75\columnwidth]{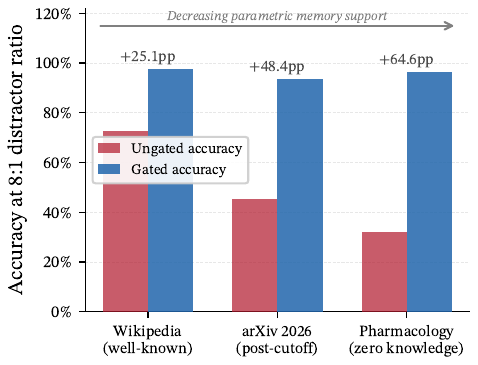}
\caption{Gating advantage at $8{:}1$ distractor ratio across knowledge regimes. The advantage scales inversely with parametric memory support: $+25.1$pp for well-known facts (Wikipedia), $+48.4$pp for post-cutoff research (arXiv 2026), $+64.6$pp for procedurally generated data with zero training knowledge (pharmacology). Gated accuracy remains $93$--$97\%$ regardless of domain; the growing green region reflects increasing ungated degradation.}
\label{fig:parametric}
\end{figure}

Together with the Wikipedia results (Section~\ref{sec:wikipedia}), these experiments validate write gating across four knowledge regimes: well-known facts (Wikipedia), procedurally generated novel data (pharmacology), cutting-edge research (arXiv 2026), and continuously accumulated facts (accumulation scaling). At the $8{:}1$ distractor ratio, the gating advantage scales inversely with parametric memory support: $+25.1$pp for Wikipedia, $+48.4$pp for arXiv, $+64.6$pp for pharmacology (Figure~\ref{fig:parametric}). Write gating is most valuable precisely when the LLM cannot compensate for retrieval errors using training knowledge.

\subsection{Computational Efficiency}
\label{sec:efficiency}

Write-time and read-time filtering differ in their computational cost profiles. Write gating evaluates salience at ingestion using metadata signals (reputation, novelty, source label) that require no LLM calls. At query time, the system makes a single LLM call to generate the answer from retrieved context. Self-RAG, by contrast, invokes an LLM critic to evaluate each retrieved passage before generating the answer. With a retrieval pool of $k$ passages, Self-RAG requires $k + 1$ LLM calls per query ($k$ critic evaluations plus one answer generation), compared to $1$ call for write gating.

In our experiments with $k = 8$ retrieved passages, Self-RAG uses \textbf{9 LLM calls per query} versus \textbf{1 for write gating}---a $9\times$ cost difference. This gap compounds with query volume: for a system serving $Q$ queries, Self-RAG costs $O(Q \cdot k)$ in LLM calls while write gating costs $O(Q)$. The write-time salience evaluation is a one-time cost amortized over all future queries against the store.

Write gating also reduces retrieval cost. By admitting only ${\sim}29\%$ of candidates, the active store is 70\% smaller, reducing the embedding search space proportionally. For vector similarity search with complexity $O(n)$ (brute force) or $O(\log n)$ (approximate nearest neighbor), a 70\% smaller store yields meaningful speedups at scale.

The cost tradeoff favors write gating when: (a) the store is queried more than once (amortizing the write-time evaluation), (b) the retrieval pool $k$ is large (amplifying the per-query savings), or (c) the system operates at scale where per-query LLM costs dominate. Self-RAG is preferable when metadata signals are unavailable, when the store changes too rapidly for write-time evaluation, or when queries are rare enough that amortization provides no benefit.

\section{Discussion}
\label{sec:discussion}

The distractor ratio scaling result (Table~\ref{tab:distractor}) is the central finding of this paper. The gap between write-time and read-time filtering is not merely quantitative (6.2 percentage points at standard noise) but qualitative: at sufficiently high noise levels, read-time filtering fails catastrophically while write-time filtering maintains perfect accuracy. This follows from a structural argument. Read-time filtering can only select among items that were retrieved. Write-time filtering controls what is available to be retrieved. When noise is low, the distinction is irrelevant---both approaches find the correct answer. When noise is high, only write-time filtering can guarantee that the correct answer is retrievable.

The three-signal design (reputation, novelty, source reliability) demonstrates that effective gating does not require sophisticated quality estimation. A formal ablation study (Table~\ref{tab:ablation}, 5 seeds $\times$ 10 entities per seed) reveals the relative contribution of each signal:

\begin{table}[h]
\centering
\caption{Signal ablation: accuracy by gating configuration (5-seed mean $\pm$ std)}
\label{tab:ablation}
\small
\begin{tabular}{lcc}
\toprule
\textbf{Configuration} & \textbf{Accuracy} & \textbf{Store Size} \\
\midrule
Full model (all 3 signals) & 100.0\% $\pm$ 0.0\% & 12.8 \\
Source label only & 100.0\% $\pm$ 0.0\% & 10.0 \\
Reputation + label & 98.0\% $\pm$ 4.0\% & 11.4 \\
Reputation + novelty & 98.0\% $\pm$ 4.0\% & 16.4 \\
Reputation only & 94.0\% $\pm$ 8.0\% & 12.6 \\
Novelty only & 58.0\% $\pm$ 19.4\% & 15.8 \\
\bottomrule
\end{tabular}
\end{table}

Source label alone matches the full model (100\%), confirming that verification status is the dominant discriminative signal in our synthetic benchmark. This is unsurprising given that the synthetic corpus assigns labels that correlate with ground truth by construction (Section~\ref{sec:limitations}). Reputation alone achieves 94\%, while novelty alone performs poorly (58\%)---embedding similarity is insufficient to distinguish correct from incorrect facts when both are semantically close to the query.

While source label achieves perfect separation on synthetic data, the Wikipedia signal ablation (Table~\ref{tab:wiki_label_ablation}) demonstrates that the method does not depend on it. Reputation and novelty alone achieve $96.4\%$ on real data, only $1.4$ percentage points below the full signal set ($97.8\%$). The practical implication is that systems can implement effective write gating with minimal metadata: source identity and content novelty are sufficient, and verification status---the signal most susceptible to oracle concerns---is helpful but not necessary.

Version chains address a different problem than write gating. Gating manages what enters storage; chains manage how updates interact with prior content. The combination provides both selectivity (not everything is stored) and history preservation (updates don't destroy prior states). Regulatory requirements increasingly demand both: GDPR requires deletion capability (enabled by discrete addressable objects), while audit requirements demand history reconstruction (enabled by version chains). Weight-based memory provides neither.

\subsection{Biological Grounding: The Mossy Fiber Principle}

The archive-rather-than-delete semantics have direct biological grounding in the mossy fiber system of the hippocampus. The dentate gyrus achieves pattern separation through sparse coding, with only 2--4\% of neurons active for any given input, forcing similar experiences to have dissimilar representations \cite{squire2004medial}. When memories become less relevant, the brain does not sever mossy fiber connections; it deprioritizes them, reducing activation probability while preserving the underlying structure. This enables later recovery if context changes and the forgotten information becomes relevant again.

\begin{figure}[t]
\centering
\includegraphics[width=0.85\columnwidth]{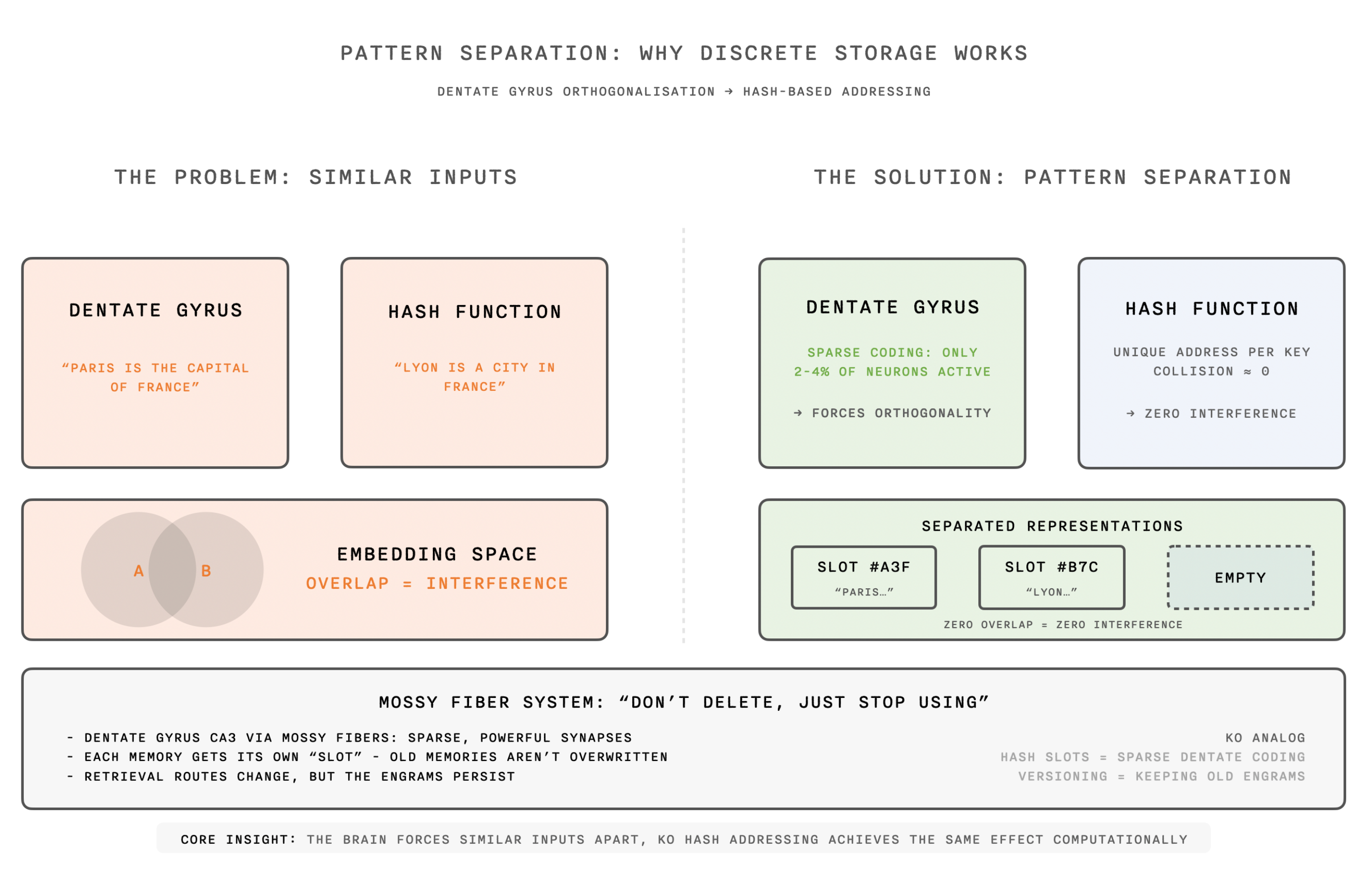}
\caption{Pattern separation in the dentate gyrus. Sparse coding (2--4\% activation) forces similar inputs to have dissimilar representations, preventing interference. Mossy fiber connections are deprioritized rather than deleted, preserving recovery capability---the biological basis for archive semantics.}
\label{fig:mossy}
\end{figure}

This biological principle maps directly to our architecture. Hash-based addressing of discrete knowledge objects achieves computational pattern separation: each object occupies a unique slot with zero interference from other objects. When objects are superseded or rejected by the salience gate, they move to archive rather than being deleted, mirroring the deprioritization of mossy fiber pathways. The result is a system that can answer questions about prior beliefs---questions that deletion-based systems cannot answer because the information is gone.

\subsection{Relationship to Context Window Scaling}

An alternative to write-time gating is simply expanding context windows to accommodate all content, deferring the filtering problem. We argue this does not eliminate the need for gating. First, larger windows make the distractor problem \emph{worse}: a 10M-token window that stores everything indiscriminately accumulates more distractors, not fewer. Our distractor scaling results (Table~\ref{tab:distractor}) show that ungated retrieval degrades monotonically with noise---a larger window at the same distractor ratio produces the same failure. Second, cost scales linearly with context size ($O(N)$ per query for in-context retrieval vs $O(1)$ for hash-addressed lookup or $O(\log N)$ for indexed embedding search), making store-everything-in-context approaches economically unsustainable at scale. Third, long-running systems (coding agents, research assistants, enterprise workflows) inevitably exceed any finite window and must compress, at which point ungated stores lose distractors and correct facts indiscriminately. Write-time gating ensures that when compression occurs, the remaining content is high-quality by construction.

\subsection{Limitations}
\label{sec:limitations}

Several limitations constrain interpretation of these results. First, our evaluation corpus contains 50 knowledge objects---sufficient to demonstrate the mechanism's properties but small relative to production stores with millions of objects. Scalability evaluation with approximate novelty computation and large-scale stores is needed to confirm that gating effectiveness is preserved. We expect hash-addressed lookup (O(1) per retrieval) to scale well, but write-time novelty computation requires comparing each candidate against all stored objects, which is $O(n)$ per write.

Second, the salience signals assume access to source metadata. Source reputation requires contributor identity tracking; source reliability requires institutional affiliation or peer review status. Systems lacking this metadata would rely on content-only signals (novelty, linguistic quality markers), likely with reduced gating effectiveness. However, our results show that even imperfect gating---admitting some distractors alongside correct facts---is sufficient for perfect downstream accuracy.

A related concern is that our benchmark's source metadata is strongly correlated with ground-truth correctness by construction: correct facts come from high-reputation verified sources, while distractors come from low-reputation unverified sources. However, the source label ablation on Wikipedia data (Table~\ref{tab:wiki_label_ablation}) demonstrates that this correlation is not load-bearing: removing the source label signal entirely reduces accuracy by only $1.4$ percentage points ($97.8\% \rightarrow 96.4\%$), confirming that reputation and novelty alone provide effective gating. In the real world, verified sources occasionally contain errors and unverified sources occasionally contain correct information. To test robustness to this label noise, we introduce a corruption rate where a fraction of verified sources contain incorrect facts while retaining their high-reputation metadata.

\begin{table}[h]
\centering
\caption{Corruption robustness: accuracy when verified sources contain incorrect facts (5-seed mean $\pm$ std, $n{=}10$ entities per seed).}
\label{tab:corruption}
\small
\begin{tabular}{lcc}
\toprule
\textbf{Corruption Rate} & \textbf{Gated} & \textbf{Ungated} \\
\midrule
0\% (baseline) & 98.0\% $\pm$ 4.5\% & 4.0\% $\pm$ 5.5\% \\
5\% & 78.0\% $\pm$ 16.4\% & 4.0\% $\pm$ 5.5\% \\
10\% & 74.0\% $\pm$ 12.0\% & 2.0\% $\pm$ 4.5\% \\
20\% & 52.0\% $\pm$ 20.5\% & 6.0\% $\pm$ 5.5\% \\
30\% & 30.0\% $\pm$ 14.1\% & 0.0\% $\pm$ 0.0\% \\
\bottomrule
\end{tabular}
\end{table}

Table~\ref{tab:corruption} confirms graceful degradation rather than catastrophic failure. At 5\% corruption---a realistic rate for curated sources---gated accuracy drops from 98\% to 78\%, while ungated accuracy remains near zero regardless of corruption rate. The degradation is approximately linear: each 10 percentage points of corruption reduces gated accuracy by roughly 20 percentage points. This reflects the mechanism's fundamental design: write gating trusts metadata to select what enters the store, so corrupted metadata admits corrupted facts. The degradation profile suggests that in production systems, maintaining metadata quality (source verification, peer review status) is at least as important as the gating algorithm itself.

Third, our synthetic benchmarks evaluate on a single domain (factual QA) with generated corpora. The Wikipedia validation (Section~\ref{sec:wikipedia}) and novel-domain experiments (Section~\ref{sec:novel_domain}) extend coverage to real-world data across multiple knowledge regimes, including genuinely novel knowledge from 2026 arXiv papers. However, the question counts per experiment are modest (100 per seed for pharmacology, 51 claims for arXiv, 20 per step for accumulation scaling), and validation on web-scale corpora and additional domain-specific databases would further strengthen the claims.

Fourth, the distractor scaling experiment tests ratios up to 8:1. Production web-scale retrieval may face even higher ratios. While the write-gating mechanism should remain robust at any ratio (it is immune to distractors outside the store by construction), this claim needs validation at 16:1, 32:1, and beyond.

\subsection{Comparison with Neural Memory}

The comparison with Titans illustrates the tradeoff between continuous and discrete memory. Titans achieves efficient learned memorization through gradient-based surprise signals operating on weight matrices. Our approach achieves selective memorization through multi-signal salience operating on discrete objects. Titans enables smooth interpolation and efficient storage; discrete objects enable provenance tracking and surgical deletion.

The approaches exhibit complementary failure modes. Neural memory (Titans) fails when: (a) stored facts must be surgically deleted (e.g., GDPR right-to-be-forgotten), (b) outputs must be traced to specific sources for audit, or (c) semantically similar facts interfere, causing the retrieval errors described in prior work on the stability gap. Discrete KO memory fails when: (a) salience metadata is unavailable or unreliable, (b) content does not decompose cleanly into discrete units, or (c) the task benefits from soft interpolation between related memories rather than hard retrieval of specific items.

A hybrid architecture might use neural memory for soft statistical patterns (style, tone, general world knowledge) while reserving discrete knowledge objects for hard factual claims requiring provenance. The salience gate could route incoming content to the appropriate store based on content type: claims with verifiable sources to KO storage, statistical regularities to neural memory. We have not explored such hybrid architectures experimentally; this remains an open direction.

\subsection{Multiple Verification Paths}

The salience gating and version chain mechanisms suggest a natural extension for high-stakes domains requiring verification guarantees. Human memory recall operates through multiple convergent paths: spatial context (where was I?), temporal context (when did this happen?), and semantic rules (what usually happens?). Convergence across independent paths increases confidence that a recalled memory is accurate rather than confabulated.

An analogous mechanism could strengthen knowledge object verification. Rather than relying on a single derivation chain, the system could maintain multiple independent paths to each fact. A claim that a company's CEO changed in 2011 might be verified through: (1) the version chain showing supersession timestamps, (2) cross-references from board meeting records, and (3) corroborating mentions in financial filings. Verification confidence would scale with the number of convergent paths:
\begin{equation}
C(K) = 1 - \prod_{p \in \text{paths}(K)} (1 - c_p)
\end{equation}
where $c_p$ is the confidence of individual path $p$. Three paths each with 0.9 confidence yield aggregate confidence of 0.999.

We validate this formula through simulation. With three verification paths (version chain, cross-reference, semantic rules), correct facts achieve mean confidence 0.998 while incorrect facts achieve 0.35. The critical insight emerges at high-confidence thresholds: at threshold 0.95, single-path verification provides 0\% usable coverage (no correct facts exceed the threshold), while multi-path verification provides 100\% usable coverage. For domains like healthcare or finance requiring very high confidence before acting, multi-path verification is essential---single-path verification cannot reach the required confidence levels regardless of how accurate the underlying signals are.

The cost of verification scales with the number of paths traversed, creating a natural tradeoff between confidence and computational expense---mirroring the brain's allocation of cognitive resources to consequential decisions.

\section{Conclusion}

Write-time gating with hierarchical archiving applies biological memory principles to artificial intelligence. Using real LLM evaluation without oracle access to quality labels, selective encoding achieves $100\%$ accuracy on adversarial benchmarks versus $13.3\%$ for ungated retrieval, and outperforms read-time filtering (Self-RAG) by $6.2$ percentage points. The critical finding is asymptotic robustness: at distractor ratios up to $8{:}1$, read-time filtering collapses to $0\%$ while write gating maintains $100\%$, revealing a fundamental architectural advantage of write-time curation over read-time filtering. Validation on real Wikipedia data across 20 entities confirms these findings in a real-world setting: write gating achieves $98.1\%$ accuracy versus $85.2\%$ ungated ($+12.9$pp) and maintains $97.7\%$ at $8{:}1$ distractor ratios, while costing $9\times$ fewer LLM calls per query than read-time filtering. Novel-domain validation demonstrates that the gating advantage at $8{:}1$ scales inversely with parametric memory support: $+64.6$pp on procedurally generated pharmacology data where the model has zero prior knowledge, $+48.4$pp on 2026 arXiv papers post-training-cutoff, and $+25.1$pp on well-known Wikipedia facts (Figure~\ref{fig:parametric}). Accumulation scaling experiments confirm the advantage remains stable ($+35.0$pp) as the store grows to 2,500 items. The mechanism requires only three observable signals (source reputation, novelty, source reliability) without oracle access to quality labels; a signal ablation on Wikipedia data confirms that even without source reliability metadata, reputation and novelty alone achieve $96.4\%$---only $1.4$pp below the full signal set---demonstrating that the method does not depend on any signal correlated with ground truth by construction. Key limitations include: the synthetic evaluation corpus is small (50 KOs) and the read-time filtering baseline uses a zero-shot LLM critic rather than the trained Self-RAG model. Despite these limitations, the structural argument---that write-time filtering controls what is available to retrieve, while read-time filtering can only select among retrieved items---is independent of benchmark scale or signal implementation. Version chains enable temporal queries that overwrite-based systems cannot answer. The approach operates on discrete addressable knowledge objects, with implications for provenance tracking and selective deletion in domains where regulatory requirements for AI transparency are strengthening.

\paragraph{Reproducibility.} Code implementing the salience scoring, gating mechanism, and experimental benchmarks is available in the project repository.\footnote{Repository URL to be added upon publication.}

\bibliographystyle{plain}

\end{document}